\RequirePackage{scrlfile}
\documentclass[hidelinks]{article}
\usepackage{subfigure}
\pdfoutput=1

\usepackage{svg}
\usepackage{parcolumns}

\usepackage{tabularx}
\usepackage{wrapfig}
\usepackage{cancel}
\usepackage{enumitem}
\PassOptionsToPackage{numbers, compress}{natbib}

\usepackage[final]{neurips_2020}
\usepackage[pdfencoding=auto, psdextra]{hyperref}

\usepackage[english]{babel}
\usepackage[utf8]{inputenc} 
\usepackage[T1]{fontenc}    
\usepackage{booktabs}
\usepackage{tabularx}

\usepackage{times}
\usepackage{hyperref}
\usepackage{url}
\usepackage{harpoon}
\usepackage{amssymb}
\usepackage{amsmath}
\usepackage{graphicx}
\usepackage{sidecap}
\usepackage{courier}
\usepackage{color}
\usepackage{caption}
\usepackage{lscape}
\usepackage{tikz}
\usepackage[ruled,vlined]{algorithm2e}
\usepackage[export]{adjustbox}
\usepackage{lscape}

\usepackage[acronym,smallcaps,nowarn,section,nogroupskip,nonumberlist]{glossaries}
\usepackage[nameinlink,capitalise]{cleveref}

\usetikzlibrary{arrows,intersections, decorations.markings}

\usepackage{amsthm}
\usepackage{booktabs}
\usepackage{nicefrac}       
\usepackage{microtype}      
\usepackage{amsfonts}       
\usepackage{harpoon}
\usepackage{parcolumns}

\newtheorem{lemma}{Lemma}

\DeclareMathOperator{\E}{\mathbb{E}}

\newcommand{\abstractmean}{{power mean }}

\newcommand{\vx}{x}
\newcommand{\vz}{z}

\newcommand{\z}{z}
\newcommand{\tpi}{\tilde{\pi}}

\DeclareMathOperator*{\argmin}{arg\,min}

\glsdisablehyper{}
\newacronym{AIS}{ais}{Annealed Importance Sampling}
\newacronym{PT}{pt}{Parallel Tempering}
\newacronym{KM}{km}{Kolmogorov-Nagumo}
\newacronym{BQ}{bq}{Bayesian Quadrature}
\newacronym{AUC}{auc}{area under the curve}
\newacronym{BAR}{bar}{Bennett's Acceptance Ratio}
\newacronym{BDMC}{bdmc}{Bidirectional Monte Carlo}
\newacronym{JS}{js}{Jensen-Shannon}
\newacronym{CFT}{cft}{Crooks's Fluctuation Theorem}
\newacronym{ELBO}{elbo}{evidence lower bound}
\newacronym{EUBO}{eubo}{evidence upper bound}
\newacronym{HMC}{hmc}{Hamiltonian Monte Carlo}
\newacronym{IB}{ib}{Information Bottleneck}
\newacronym{IS}{is}{importance sampling}
\newacronym{IWAE}{iwae}{importance weighted autoencoder}
\newacronym{KL}{kl}{Kullback Leibler}
\newacronym{MCMC}{mcmc}{Markov Chain Monte Carlo}
\newacronym{RD}{r-d}{rate-distortion}
\newacronym{RWS}{rws}{reweighted wake-sleep}
\newacronym{SGD}{sgd}{stochastic gradient descent}
\newacronym{SNIS}{snis}{self-normalized importance sampling}
\newacronym{TI}{ti}{thermodynamic integration}
\newacronym{TVI}{tvi}{thermodynamic variational inference}
\newacronym{TVO}{tvo}{thermodynamic variational objective}
\newacronym{VAE}{vae}{Variational autoencoders}
\newacronym{VI}{vi}{variational inference}
\newacronym{VIMCO}{vimco}{variational inference for Monte Carlo objectives}
\newacronym{WS}{ws}{wake-sleep}

\title{Annealed Importance Sampling with q-Paths}

\author{
Rob Brekelmans$^1$\thanks{equal contribution}~,
Vaden Masrani$^2$\footnotemark[1]~,
Thang Bui$^3$,\\
{\bf Frank Wood$^2$,
Aram Galstyan$^1$,
Greg Ver Steeg$^1$, 
Frank Nielsen$^4$
}\\ 
$^1$USC Information Sciences Institute, $^2$University of British Columbia, $^3$UberAI, $^4$Sony CSL \\
{\tt \{brekelma,galstyan,gregv\}@isi.edu},
{\tt \{vadmas,fwood\}@cs.ubc.ca},\\
{\tt thang.bui@uber.com},
{\tt frank.nielsen@acm.org }
}

\begin{document}

\maketitle
\begin{abstract}
\gls{AIS} \cite{neal2001annealed, jarzynski1997equilibrium} is the gold standard for estimating partition functions or marginal likelihoods, corresponding to importance sampling over a path of distributions between a tractable base and an unnormalized target.  While \gls{AIS} yields an unbiased estimator for \textit{any} path, existing literature has been primarily limited to the geometric mixture or moment-averaged paths associated with the exponential family and \textsc{kl} divergence \cite{grosse2013annealing}.   We explore \gls{AIS} using $q$-paths, which include the geometric path as a special case and are related to the homogeneous power mean, deformed exponential family, and $\alpha$-divergence \cite{amari2007integration}.
\end{abstract}









 %


\section{Introduction}
\gls{AIS} \cite{neal2001annealed, jarzynski1997equilibrium} is a method for estimating intractable normalization constants, which considers a path of intermediate distributions $\pi_t(z)$ between a tractable base distribution $\pi_0(\vz)$ and unnormalized target $\tpi_T(\vz)$.  
In particular, \gls{AIS} samples from a sequence of \textsc{mcmc} transition operators $\mathcal{T}_{t}(\vz_{t}|\vz_{t-1})$ which leave each $\pi_{\beta_t}(\vz) = \tpi_{\beta_t}(\vz)/ Z_{t} $ invariant to estimate the ratio $Z_T / Z_0$.
\begin{wrapfigure}{r}{0.35\textwidth}
    \begin{minipage}{0.35\textwidth}
            \vspace{-5pt}
            \begin{algorithm}[H]
                \SetAlgoLined
                    \For{i = 1 \text{to} N}{
                    $z_0 \sim \pi_0(z)$\\
                    $w^{(i)} \leftarrow Z_0$\\
                    \For{t = 1 to T}{
                    $w_t^{(i)} \leftarrow$ $w_t^{(i)} \frac{\tpi_t(z_{t - 1}^{(i)})}{\tpi_{t-1}(z_{t - 1}^{(i)})}$\\
                    $z_t^{(i)} \sim T_t(\vz_t | \vz_{t-1}^{(i)})$\\ 
                    }
                    }
                    \Return{$Z_T / Z_0 \approx \frac{1}{N} \sum_N w_T^{(i)}$}
                    \caption{Annealed IS}
                    \label{alg:ais}
            \end{algorithm}
        \end{minipage}
        \vspace{-14pt}
\end{wrapfigure}
As shown in \cref{alg:ais}, we can accumulate the importance weights $w_{T}^{(i)} = \prod_{t=1}^{T} \tpi_t(z_{t - 1}) / \tpi_{t-1}(z_{t - 1})$ along the path.   Taking the expectation of $w_{T}^{(i)}$ over sampling chains yields an unbiased estimate of $Z_T  / Z_0$ \cite{neal2001annealed}.
Similarly, \gls{BDMC} \cite{grosse2015sandwiching, grosse2016measuring} provides lower and upper bounds on the \textit{log} partition function ratio $\log Z_T / Z_0$ using \gls{AIS} initialized with the base or target distribution, respectively.

\gls{AIS} often uses a geometric mixture path with schedule $\{ \beta_t \}_{t=0}^T$  to anneal between $\pi_0$ and $\pi_T$, 
\begin{align}
    \tilde{\pi}_{\beta}(\vz) & = \tpi_0(\vz)^{1-\beta} \, \tpi_T(\vz)^{\beta}, \label{eq:geopath}
\end{align}
where $\pi_{\beta}(\vz) = \tilde{\pi}_{\beta}(\vz) / Z_{\beta}$ and $Z_{\beta} = \int \tpi_0(\vz)^{1-\beta} \tpi_T(\vz)^{\beta} d\vz$. 

Alternative paths have been discussed in \cite{grosse2013annealing,gelman1998simulating, thang}, but may not have closed form expressions for intermediate distributions. 
In this work, we propose to generalize the geometric mixture path \eqref{eq:geopath} using the \abstractmean \cite{kolmogorov1930,hardy1953, de2016mean}, or $q$-path,
\begin{align}
    \tpi^{(q)}_{\beta}(\vz) &= \bigg[ (1-\beta) \, \tpi_0(\vz)^{1-q} + \beta \, \tpi_T(\vz)^{1-q} \bigg]^{\frac{1}{1-q}}\label{eq:alpha_mix0}
\end{align}
As $q\rightarrow 1$, we recover the geometric mixture path as a special case. The \abstractmean 
is derived using the \textit{q-logarithm} function from non-extensive thermodynamics \cite{tsallis1988possible, naudts2011generalised, tsallis2009introduction}, which allows us to frame \cref{eq:alpha_mix0} in terms of the the $q$-exponential family \cite{amari2011q}. 
Further, we draw connections with the $\alpha$-integration of \citet{amari2007integration,amari2016information} by showing that \cref{eq:alpha_mix0} minimizes a mixture of $\alpha$-divergences as in \cite{amari2007integration}. 
We describe properties of the geometric and $q$-paths in \cref{sec:alternative_interpretations} and \cref{sec:alpha_paths}, respectively.

\section{Interpretations of the Geometric Path}\label{sec:alternative_interpretations}
We give three complementary interpretations of the geometric path defined in \cref{eq:geopath}, which will have generalized analogues in \cref{sec:alpha_paths}.
\paragraph{Log Mixture}
Simply taking the logarithm of both sides of the geometric mixture \eqref{eq:geopath} shows that $\tpi_\beta$ can be obtained by taking the log-mixture of $\tpi_0$ and $\tpi_T$ with mixing parameter $\beta$,
\begin{align}
\log \tpi_\beta (\vz) = (1-\beta) \, \log \tpi_0(\vz) + \beta \, \log \tpi_T(\vz) \label{eq:log_mixture}
\end{align}
where we may also choose to subtract a constant $\log Z_\beta$ to enforce normalization.
\paragraph{Exponential Family} Distributions along the geometric path may also be viewed as coming from an exponential family \cite{brekelmans2020tvo, grunwald2007minimum}. In particular, we use a base measure of $\tpi_0(\vz)$ and sufficient statistics $\phi(\vz) = \log \tpi_T / \tpi_0$ to rewrite \cref{eq:geopath} as
\begin{align}
    \pi_{\beta}(\vz) &= \tpi_0(\vz) \, \exp \{ \beta \cdot \phi(\vz) - \psi(\beta) \}\label{eq:exp_fam}
\end{align}
where the mixing parameter $\beta$ appears as the natural parameter of the exponential family and $\psi(\beta) := \log Z_\beta$.
The log-partition function or free energy $\psi(\beta)$ is convex in $\beta$ and induces \cite{amari2016information, nielsen2018elementary, brekelmans2020tvo} a Bregman divergence over the natural parameter space equivalent to the \textsc{kl} divergence $D_{KL}[\pi_{\beta'}||\pi_{\beta}]$.


\paragraph{Variational Representation} 
\citet{grosse2013annealing} also observe that each $\pi_{\beta}(\vz)$ can be viewed as minimizing a weighted sum of \textsc{kl} divergences to the (normalized) base and target distributions 
\begin{align}
    \pi_{\beta}(\vz) &= \argmin \limits_{r(\vz)} \, (1-\beta) \, D_{KL}[ r(\z) || \pi_{0}(\vz)] + \beta D_{KL}[  r(\z) || \pi_{T}(\vz)] \label{eq:vrep}.
\end{align}
While the optimization in \cref{eq:vrep} is over arbitrary $r(\vz)$, the optimal solution is the geometric mixture with mixing parameter $\beta$, which is a member of the exponential family in \cref{eq:exp_fam}\cite{grosse2013annealing, brekelmans2020tvo}.

\section{Interpretations of the $q$-Path}\label{sec:alpha_paths}
To anneal between $\tpi_0$ and $\tpi_T$, we consider the power mean with order parameter $q$  in place of the geometric average in \cref{eq:geopath}.  Analogously to Sec. \ref{sec:alternative_interpretations} above, our generalization is associated with the deformed log mixture, $q$-exponential family, and a variational representation using the $\alpha$-divergence.

\paragraph{Power Means} \citet{kolmogorov1930} proposed a generalized notion of the mean using any monotonic function $h(u)$, with $h(u) = u$ corresponding to the arithmetic mean and 
\begin{align}
\mu_{h}(\{w_i, u_i\}) = h^{-1} \bigg(\sum \limits_{i} w_i \, h(u_i)\bigg), \label{eq:abstract_mean}
\end{align}
where $\mu_{h}$ outputs a scalar given a normalized measure $\{ w_i \}$ over a set of elements $\{ u_i \}$ \cite{de2016mean}.  The geometric and arithmetic means are \textit{homogeneous}, meaning they have the linear scale-free property ${\mu_{h}(\{w_i , c \cdot u_i \}) = c \cdot \mu_{h}(\{w_i ,  u_i \})}$. In order for a generalized mean to be homogenous, \citet{hardy1953} (pg. 68 or \cite{amari2007integration}) show that $h(u)$ must be of the form
\begin{align}
h_{q}(u) =
\begin{cases}
    a \cdot u^{1-q} + b & q \neq 1 \\
    \log u \hfill 	&  q = 1
\end{cases}. \label{eq:alpha_abstract}
\end{align}
which we refer to as the $q$-power mean.  Notable examples of the \abstractmean include the arithmetic mean at $q = 0$, geometric mean as $q \rightarrow 1$, and the $\min$ or $\max$ operation as $q \rightarrow \pm \infty$.  For $q = \frac{1+\alpha}{2}$, $h_{q}(u)$ matches the $\alpha$-representation of \citet{amari2016information}\cite{ amari2010divergencefunctions, amari2007methods}.

Using the \abstractmean to generalize geometric mean, we propose the $q$-path of intermediate unnormalized densities $\tilde{\pi}^{(q)}_\beta(\vz)$ for \gls{AIS}.  In App. \ref{app:any_h}, we show that for any choice of $a$ and $b$, $h_{q}(u)$ yields the same \abstractmean
\begin{align}
    \tilde{\pi}^{(q)}_\beta(\vz) = \begin{cases}\big[ (1-\beta) \, \tpi_0(\vz)^{1-q} + \beta \, \tpi_T(\vz)^{1-q} \big]^{\frac{1}{1-q}} & q \neq 1 \\[1.3ex]
     \exp \big\{ (1-\beta) \, \log \tpi_0(\vz) + \beta \, \log \tpi_T(\vz) \big\} & q = 1  \\[1.25ex]
    \end{cases},  \label{eq:alpha_mix}
\end{align}
where we have chosen $\{ w_i \} = \{ 1-\beta , \beta \}$ and $\{ u_i \} = \{ \tpi_0, \tpi_T \} $ in \eqref{eq:abstract_mean}.  

\paragraph{Deformed Log Mixture}
The deformed, or $q$-logarithm \cite{naudts2011generalised}, which plays a crucial role in non-extensive thermodynamics \cite{tsallis1988possible, tsallis2009introduction}, is a particular special case of $h_q(u)$ in \cref{eq:alpha_abstract}, with
\begin{align}
    &\ln_q(u) = \frac{1}{1-q} \big( u^{1-q} - 1 \big) & &\exp_q(u) = \big[ 1 + (1-q) \, u \big]_{+}^{\frac{1}{1-q}}, \label{eq:lnqexpq}
\end{align}
where we have also defined the $q$-exponential with $\exp_q(u) = \ln_q^{-1}(u) $ and $[ x]_{+} = \max\{0, x\}$  ensuring  $g(u)$ is non negative.   Note that $\lim_{q \to 1} \ln_q(u) = \log u$ and $\lim_{q \to 1} \exp_q(u) = \exp u$. 

Applying $h_{q}(u) =\ln_q(u)$ to both sides of \cref{eq:abstract_mean} or \eqref{eq:alpha_mix}, we can write $\tilde{\pi}^{(q)}_{\beta}$ as a deformed log-mixture
\begin{align}
    \ln_q \tilde{\pi}^{(q)}_{\beta}(\vz)  = (1-\beta) \, \ln_q \tpi_0 (\vz) \,  + \beta \, \ln_q \tpi_T (\vz) \label{eq:ln_q_mix}
\end{align}
with mixing weight $\beta$. We also provide detailed derivations for \cref{eq:ln_q_mix} in App. \ref{app:ln_qmix}.




\begin{figure}[t]
    \centering
    \subfigure[$q=0$]{\includegraphics[trim={0 1.2cm 0 1cm},clip,width=0.24\textwidth]{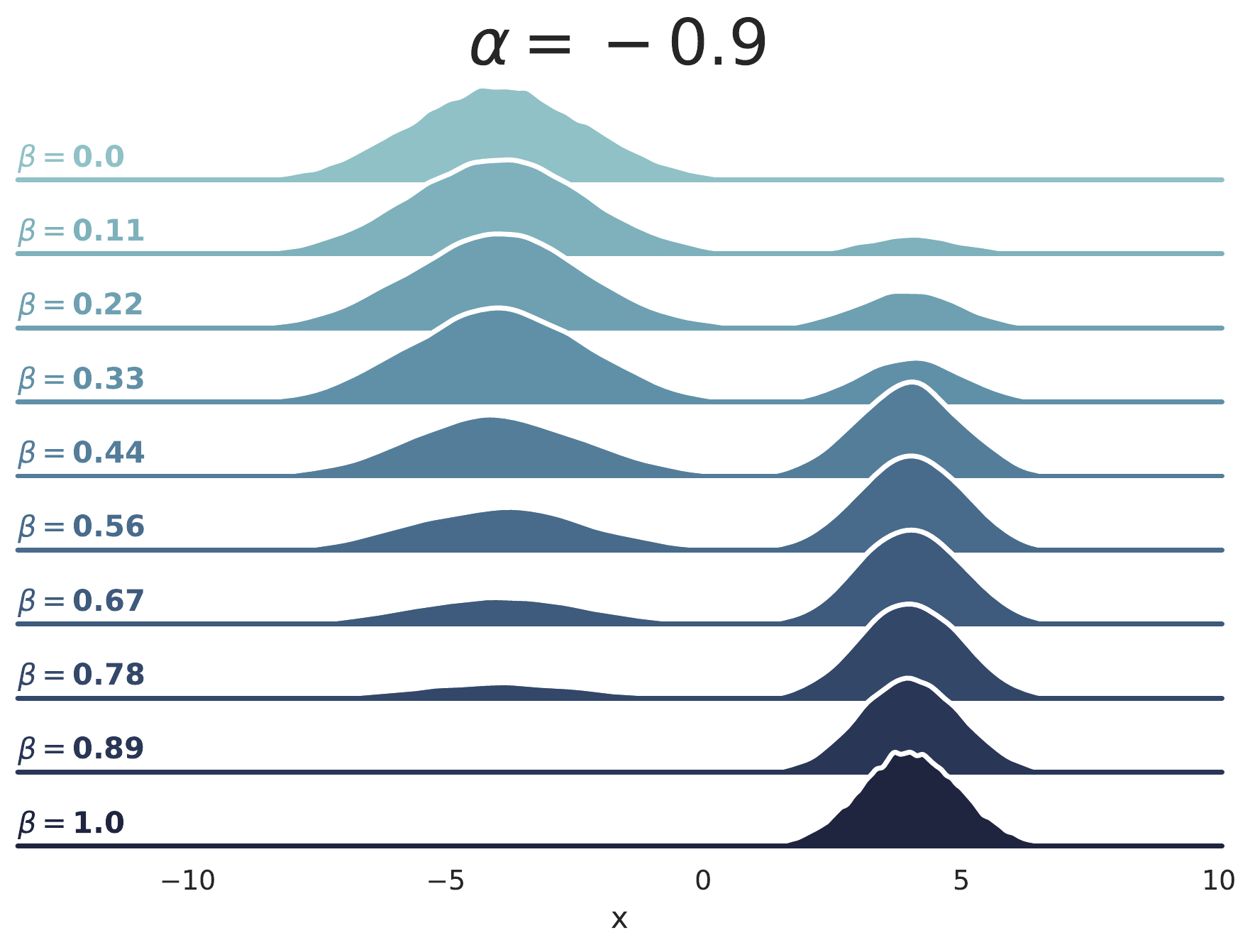}}
    \subfigure[$q=0.5$]{\includegraphics[trim={0 1.2cm 0 1cm},clip,width=0.24\textwidth]{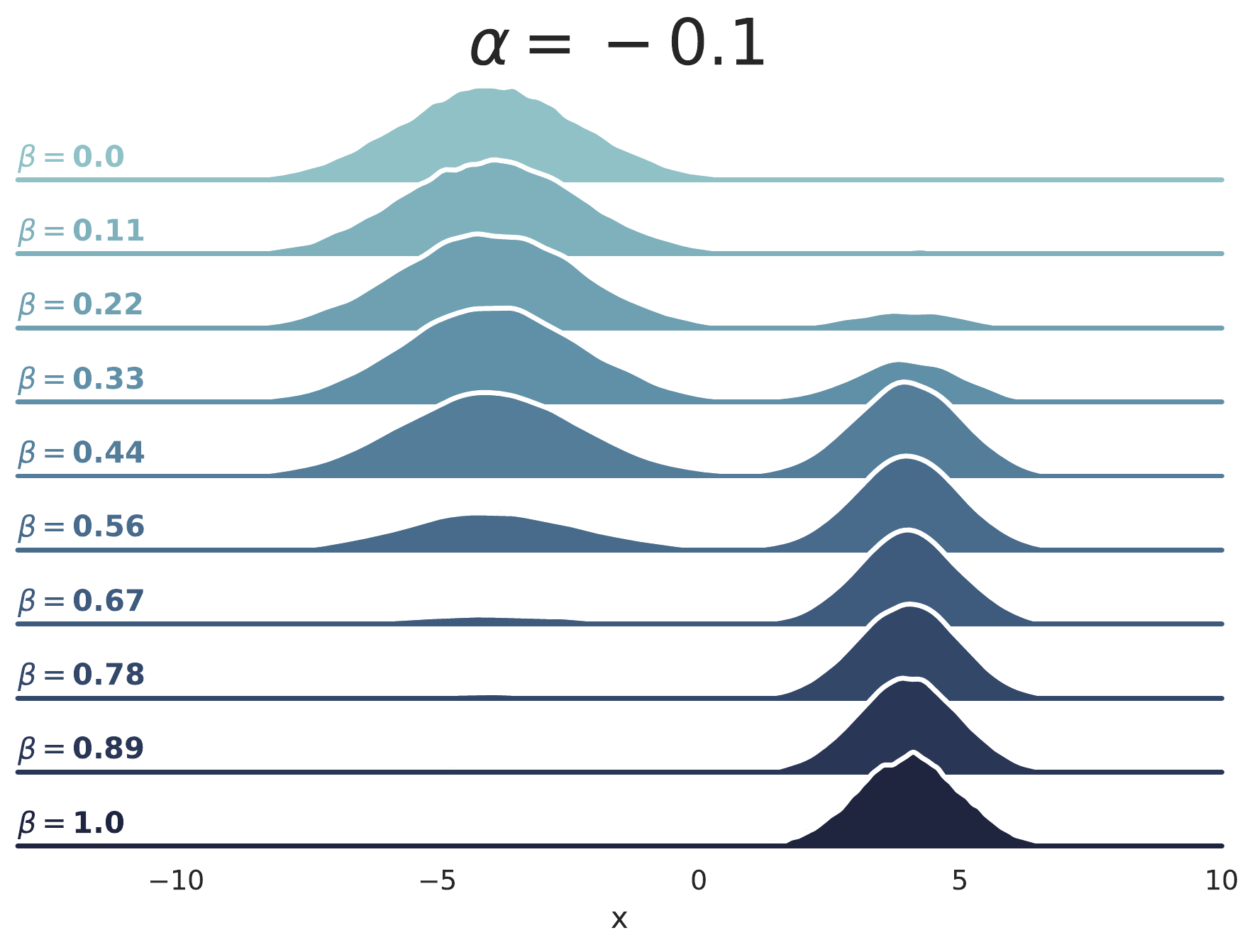}}
    \subfigure[$q=0.9$]{\includegraphics[trim={0 1.2cm 0 1cm},clip,width=0.24\textwidth]{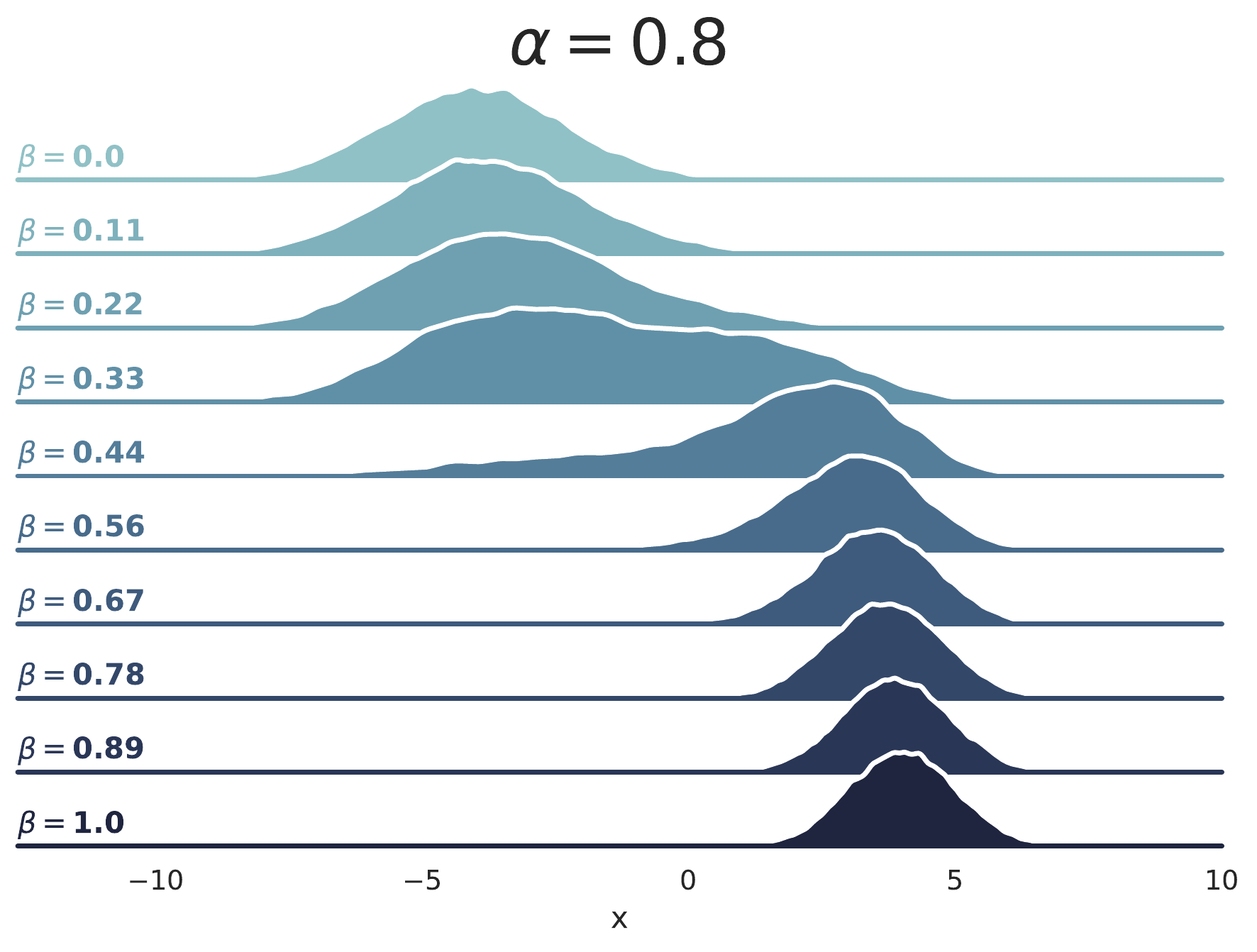}}
    \subfigure[$q=1$]{\includegraphics[trim={0 1.2cm 0 1cm},clip,width=0.24\textwidth]{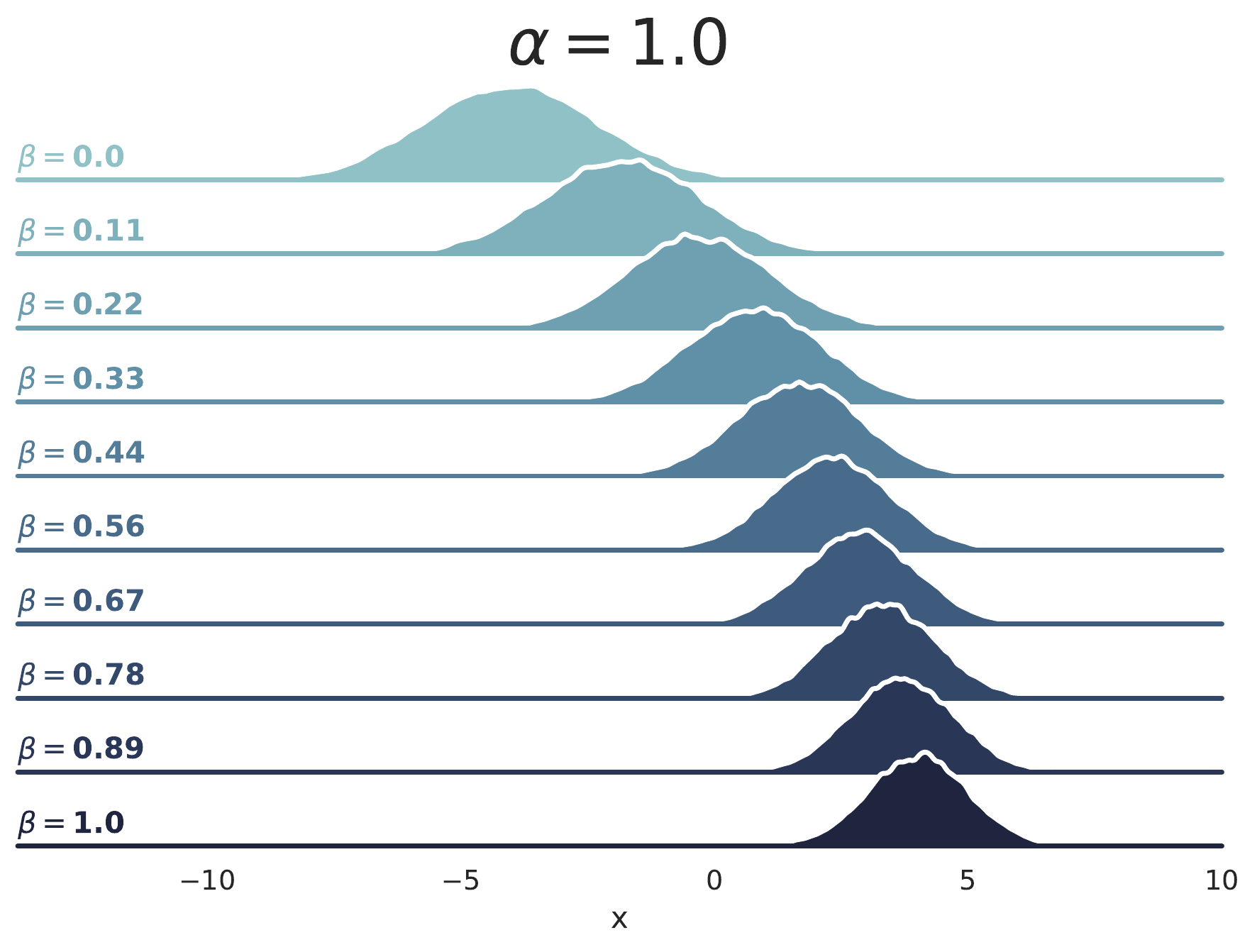}}
    \caption{Intermediate densities between $\mathcal{N}(-4, 3)$ and $\mathcal{N}(4,1)$ for various $q$-paths and 10 equally spaced $\beta$. The path approaches a mixture of Gaussians with  weight $\beta$ at $q=0$.  For the geometric mixture ($q=1$), intermediate $\pi_{\beta}$ stay within the exponential family since both $\pi_0$, $\pi_T$ are Gaussian.}
    \label{fig:alpha_path} 
    \vspace*{-.15cm}
\end{figure}

\paragraph{$q$-Exponential Family}
The $q$-exponential in \cref{eq:lnqexpq} may be used to define a  $q$-exponential family of distributions \cite{amari2011q, naudts2011generalised}.  Using $\theta$ as the natural parameter,  
\begin{align}
    \pi_{\theta}^{(q)}(\vz)  &= \tpi_0(z) \, \exp_q \big\{ \theta \cdot \phi_q(\vz) - \psi_q(\theta ) \big\} \, , \label{eq:qexp_fam} 
\end{align}
which recovers the standard exponential family at $q \to 1$. In App. \ref{app:qexp_derivation} we show that the $q$-mixture $\tilde{\pi}^{(\alpha)}_\beta$ in \cref{eq:alpha_mix} can be rewritten in terms of the $q$-exponential family 
\begin{align}
    \pi_{\beta}^{(q)}(\vz) &= \frac{1}{Z^{(q)}_{\beta}} \, \tpi_0(\z) \, \exp_q \big\{ \beta \cdot  \ln_q \frac{\tpi_T(\vz)}{\tpi_0(\vz)} \big \} 
    & & Z^{(q)}_{\beta} = \int \tpi_{\beta}^{(q)}(\z) \, d\vz \label{eq:z_norm}
\end{align}
   with sufficient statistic $\phi_q(\vz) = \ln_q \tpi_T / \tpi_0$ and natural parameter $\beta$.  The expression in \eqref{eq:z_norm} might be used to directly estimate the normalization constant $Z^{(q)}_{\beta}$ via Monte Carlo approximation.
   
   
   
As for the standard exponential family, the $q$-free energy $\psi_q(\theta)$ in \cref{eq:qexp_fam} is convex in $\theta$ and can be used to construct a Bregman divergence over normalized $q$-exponential family distributions \cite{amari2011q}.  
However, to normalize \eqref{eq:z_norm} using the $q$-free energy, a non-linear mapping $\theta(\beta)$ between parameterizations is required. This delicate issue of normalization in the $q$-exponential family has been noted in \cite{matsuzoe2019normalization,suyari2020advantages,naudts2011generalised}, and we provide more detailed discussion in App. \ref{app:normalization}.

\paragraph{Variational Representation using the $\alpha$-Divergence}
Since we do not have access to normalization constants in the \gls{AIS} setting, we focus on the $\alpha$-divergence \cite{amari1982differential, amari2016information} over unnormalized measures $\tilde{q}(\vz)$ and $\tilde{p}(\vz)$.  
We first recall the definition, 
\begin{align}
\hspace*{-.2cm} D_{\alpha}[\tilde{q}(\vz):\tilde{p}(\vz) ] &= \frac{4}{(1-\alpha^2)} \bigg( \frac{1-\alpha}{2}  \int \tilde{q}(\vz) \,d\vz +  \frac{1+\alpha}{2}  \int \tilde{p}(\vz) \,d\vz -\int \tilde{q}(\vz)^{\frac{1-\alpha}{2}} \, \tilde{p}(\vz)^{\frac{1+\alpha}{2}}  d\vz  \bigg) \nonumber  
\end{align}
which is an $f$-divergence \cite{ali1966general} for the generator $f(u) = \frac{4}{1-\alpha^2}\big( \frac{1-\alpha}{2} + \frac{1+\alpha}{2} u -  u^{\frac{1+\alpha}{2}} \big)$ \cite{amari2016information, amari2010divergencefunctions}.  Note that $\lim \limits_{\alpha \rightarrow 1} D_{\alpha}[\tilde{q}(\vz):\tilde{p}(\vz) ] = D_{KL}[ \tilde{p}(\vz):\tilde{q}(\vz) ]$ and $\lim \limits_{\alpha \rightarrow -1} D_{\alpha}[\tilde{q}(\vz):\tilde{p}(\vz) ] = D_{KL}[\tilde{q}(\vz):\tilde{p}(\vz) ]$.
\footnote{ We extend to unnormalized measures using $D_{KL}[\tilde{q}(\vz):\tilde{p}(\vz) ] = D_{KL}[q(\vz):p(\vz) ] - \int \tilde{q}(\vz) d\vz +  \int \tilde{p}(\vz) d\vz $.}


 In App. \ref{app:alpha_integration}, we follow similar derivations as \citet{amari2007integration} to show that, for $q = \frac{1+\alpha}{2}$ (\cite{amari2016information} Ch. 4), the $q$-path density $\tpi^{(q)}_{\beta}$ minimizes the expected $\alpha$-divergence to the endpoints 
\begin{align}
\tpi^{(q)}_{\beta}(\vz) &= \argmin \limits_{\tilde{r}(\vz)} \, (1- \beta) \, D_{\alpha}[ \tpi_0 (\vz) : \tilde{r}(\vz)  ] + \, \beta \, D_{\alpha}[ \tpi_T (\vz) : \tilde{r}(\vz)  ] \label{eq:vrep_alpha} \, ,
\end{align}
where the optimization is over arbitrary $\tilde{r}(\vz)$.  
This  variational representation  generalizes \cref{eq:vrep}, since the \textsc{kl} divergence is recovered (with the order of the arguments reversed) as $\alpha \rightarrow 1$ or $q \rightarrow 1$.





\paragraph{Moment-Matching Procedures}
At $q=0$, the solution to the optimization \eqref{eq:vrep_alpha} correponds to the arithmetic mean, or mixture distribution $\tpi_{t}^{(0)}(\vz) = (1-\beta) \, \tpi_0 + \beta \, \tpi_1$. 
 While the `moment-averaged' \gls{AIS} path \cite{grosse2013annealing} appears related to the $q=0$ case, we clarify in App. \ref{app:mixture_path} that  \citet{grosse2013annealing} restrict to optimization within an exponential family of distributions.  Generalizing this approach to the $\alpha$-divergence, \citet{thang} follows \citet{minka2005divergence} (Sec. 3.1-2) to derive the moment-matching condition
\begin{align}
\tilde{r}^{*}_{t, \alpha}(\vz) &:= \argmin \limits_{\tilde{r}(\vz)} (1-\beta) \, D_{\alpha}[ \tpi_0 (\vz) : \tilde{r}(\vz)  ] + \, \beta \, D_{\alpha}[ \tpi_T (\vz) : \tilde{r}(\vz)  ] \label{eq:vrep_alpha2} \\
\implies \mathbb{E}_{\tilde{r}_{*}}[ \phi(\vz) ] &= (1-\beta) \mathbb{E}_{\tpi_0^{\alpha} r_{*}^{1-\alpha} }[ \phi(\vz) ] + \beta \, \mathbb{E}_{\tpi_T^{\alpha} \tilde{r}_{*}^{1-\alpha} }[ \phi(\vz) ] \label{eq:moment_matching} 
\end{align}
where $\tilde{r}(\vz)$ comes from an exponential family  with sufficient statistics $\phi(\vz)$. 

However, we note that our $q$-path is more general than these approaches, since the optimization in \cref{eq:vrep_alpha} is over all unnormalized distributions.  Unlike the moment matching conditions above, our closed form expression for $\tpi_{\beta}^{(q)}$ can be directly used as an energy function for \textsc{mcmc} sampling.

\begin{figure*}[t]
    \begin{minipage}{.60\textwidth}
        \begin{center}
            \includegraphics[trim={0 0.4cm 0 0cm},clip,width=.915\textwidth]{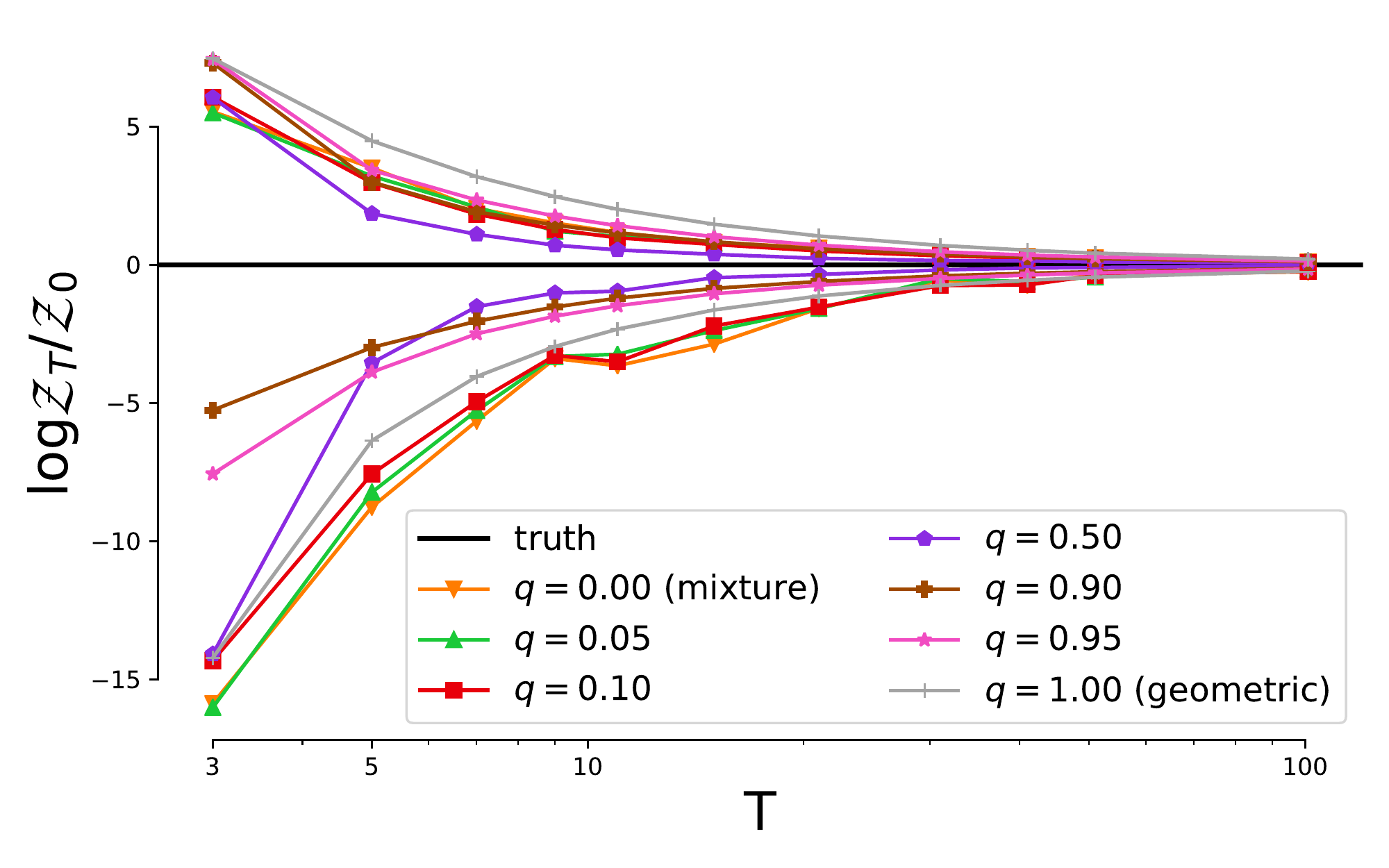}
            \vspace{-10pt}
        \captionof{figure}{\gls{BDMC} lower and upper bound estimates of $\log Z_T/Z_0$ by $q$-path order and number of intermediate distributions ($T$), for annealing between $\mathcal{N}(-4,3) \to \mathcal{N}(4,1)$.
        }\label{fig:bdmc}
        \end{center}
        \vspace{-10pt}
    \end{minipage}\hspace*{.05\textwidth}
    \begin{minipage}{.34\textwidth}
        \begin{tabular}{ll}
            \toprule
            $q$    &  $Z_{\text{est}}$ ($Z_{\text{true}} = 1$) \\
            \midrule
            0.00 (mix) &  1.0136 $\pm$ 0.0634\\
            0.05 &  1.0105 $\pm$ 0.0569 \\
            0.10 &  1.0198 $\pm$ 0.0576 \\
            \textbf{0.90} &  \textbf{0.9975} $\pm$ \textbf{0.0085} \\
            0.95 &  0.9971 $\pm$ 0.0092 \\
            1.00 (geo) &  0.9967 $\pm$ 0.0094 \\
            \bottomrule
        \end{tabular}
        \vspace{-2pt}
        \captionof{table}{ Partition Function Estimates for various $q$ and linearly spaced ${T=100}$.   A path with $q=0.90$ outperforms both the mixture of Gaussians ($q=0$) and geometric ($q=1$) paths in terms of $Z_{\text{err}} = |Z_{\text{est}} - Z_{\text{true}}|$. }\label{tab:logzb}
        \vspace{-10pt}
    \end{minipage}
\end{figure*}

\section{Experiments}

We consider $q$-paths between $\pi_0=\mathcal{N}(-4,3)$ and $\pi_T = \mathcal{N}(4,1)$ to estimate $Z_T/Z_0 = 1$, and use 
parallel runs of \gls{HMC} \cite{neal2011mcmc} to obtain accurate, independent samples from $\tilde{\pi}^{(q)}_t(\vz)$ linearly spaced between $\beta_0=0$ and $\beta_T=1$.  For all experiments, we use 10k samples from each intermediate distribution and average results across 20 seeds.  

In \cref{fig:bdmc}, we report \gls{BDMC} upper and lower bound estimates of $\log Z_T/Z_1$ for various $q$ and $T$.  We observe that the choice of $q$ can impact performance, with $q = 0.9$ obtaining tighter estimates at small $T$ and $q=0.5$ converging more quickly as $T$ increases.  Both outperform the baseline geometric path at $q=1$.  In \cref{tab:logzb}, we estimate $Z_T/Z_0$ using \gls{AIS} for $T=100$, and observe that our the $q=0.9$ path can achieve a lower error than the geometric path.

Finally, in App. \ref{app:additional}, we provide additional analysis for annealing between two Student-$t$ distributions.   The Student-$t$ family can be shown to correspond to a $q$-exponential family \cite{matsuzoe2015deformed}, with the same sufficient statistics as a Gaussian, and a degrees of freedom parameter $\nu$ that induces heavier tails and sets the value of $q$.  As $q\rightarrow 1$ or $\nu \rightarrow \infty$, the standard Gaussian is recovered.  In Fig. \ref{fig:gaussian_path}-\ref{fig:student_path}, we compare annealing between two Student-$t$ distributions in the $q=2$ family to the Gaussian case of $q=1$, and observe that the same $q$-path can induce different qualitative behavior based on properties of the endpoint distributions.

\section{Conclusion}

In this work, we propose $q$-paths to generalize the geometric mixture path commonly used in \gls{AIS}, and show that modifying the path can improve \gls{AIS} and \gls{BDMC} for a fixed mixing schedule on a toy Gaussian example.   We interpreted our $q$-paths using the deformed logarithm, $q$-exponential family, and $\alpha$-divergences, which may suggest further connections in non-extensive thermodynamics and information geometry.  
Choosing a schedule for a given $q$-path, understanding how the choice of $q$ depends on properties of the initial and target distributions, and exploring the use of $q$-paths in related methods such as the \gls{TVO} \cite{masrani2019thermodynamic, brekelmans2020tvo} remain interesting directions for future work. 

\bibliographystyle{plainnat}
\bibliography{ref}
\clearpage
\appendix
\section{Abstract Mean is Invariant to Affine Transformations} \label{app:any_h}
In this section, we show that $h_{q}(u)$ is invariant to affine transformations. That is, for any choice of $a$ and $b$,
\begin{align}
    h_{q}(u) =
\begin{cases}
    a \cdot u^{1-q} + b \hfill & q \neq 1 \\
    \log u \hfill 	&  q = 1
\end{cases} \label{eq:alpha_abstract2}
\end{align}
yields the same expression for the abstract mean $\mu_{h_{\alpha}}$. First, we note the expression for the inverse $h^{-1}_{q}(u)$ at $q \neq 1$
\begin{align}
    h^{-1}_{q}(u) = \left(\frac{u - b}{a}\right)^{\frac{1}{1-q}}.
\end{align}
Recalling that $\sum_i w_i = 1$, the abstract mean then becomes
\begin{align}
    \mu_{h_{q}}(\{w_i\}, \{u_i\}) &= h_{q}^{-1}\left(\sum_i w_i h_{q}(u_i) \right) \\
    &= h_{q}^{-1}\left(a \left(\sum_i w_iu_i^{1-q}\right) + b \right) \\
    &=\bigg(\sum_i w_i u_i^{1-q} \bigg)^{\frac{1}{1-q}}
\end{align}
which is independent of both $a$ and $b$. 

\section{Derivations of the $q$-Path}\label{app:q_path}
\subsection{Deformed Log Mixture}\label{app:ln_qmix}
In this section, we show that the unnormalized $\ln_q$ mixture
\begin{align}
\ln_q \tpi^{(q)}_{\beta}(\z) &=  (1-\beta) \ln_q \tpi_0(\z) + \beta \ln_q \tpi_1(\z)
\end{align}
reduces to the form of the $q$-path intermediate distribution in \eqref{eq:alpha_mix0} and \eqref{eq:alpha_mix}.  Taking $\exp_q$ of both sides,
\begin{align*}
    \tpi^{(q)}_{\beta}(\z) &=  \exp_q \left\{ (1-\beta) \ln_q \tpi_0(\z) + \beta \ln_q \tpi_1(\z)  \right\} \\
    &= \left[ 1 + (1-q) \left( \ln_q \tpi_0(\z) + \beta \left( \ln_q \tpi_1(\z) - \ln_q \tpi_0(\z) \right)\right) \right]_{+}^{\frac{1}{1-q}} \\
    &= \left[ 1 + (1-q) \frac{1}{1-q} \left( \tpi_0(\z)^{1-q} - 1  + \beta \big( \tpi_1(\z)^{1-q} - 1 - \tpi_0(\z)^{1-q} + 1 \big)\right)   \right]_{+}^{\frac{1}{1-q}} \\
  &= \left[ 1 +  \tpi_0(\z)^{1-q} - 1  + \beta \bigg( \tpi_1(\z)^{1-q}  - \tpi_0(\z)^{1-q} \bigg) \right]_{+}^{\frac{1}{1-q}} \\
  &= \left[  \tpi_0(\z)^{1-q}  + \beta \, \tpi_1(\z)^{1-q} - \beta \, \tpi_0(\z)^{1-q}    \right]_{+}^{\frac{1}{1-q}} \\
  &= \left[  (1-\beta ) \, \tpi_0(\z)^{1-q}  + \beta \, \tpi_1(\z)^{1-q}  \, \right]_{+}^{\frac{1}{1-q}}
\end{align*}

\subsection{$q$-Exponential Family}\label{app:qexp_derivation}
Here, we show that the unnormalized $q$-path reduces to a form of the $q$-exponential family
\begin{align}
    \tpi^{(q)}_{\beta}(\vz) &= \bigg[(1 - \beta) \tpi_0(\vz)^{1 - q} + \beta \tpi_1(\vz)^{1 - q}\bigg]^\frac{1}{1 - q}\\
    &= \bigg[\tpi_0(\vz)^{1 - q} + \beta\big(\tpi_1(\vz)^{1 - q} - \tpi_0(\vz)^{1 - q}\big)\bigg]^\frac{1}{1 - q}\\
    &= \tpi_0(\vz) \left[1 + \beta\left(\left(\frac{\tpi_1(\vz)}{\tpi_0(\vz)}\right)^{1 - q} - 1\right)\right]^\frac{1}{1 - q}\\
    &= \tpi_0(\vz) \left[1 + (1-q) \, \beta \, \ln_q\left( \frac{\tpi_1(\vz)}{\tpi_0(\vz)}\right)\right]^\frac{1}{1 - q}\\
    &= \tpi_0(\vz) \exp_q \left\{ \beta \cdot \ln_q\left(\frac{\tpi_1(\vz)}{\tpi_0(\vz)}\right)\right\}.
\end{align}
Defining $\phi(\vz) = \ln_{q} \frac{\tpi_1(\vz)}{\tpi_0(\z)}$ and introducing a multiplicative normalization factor $Z_q(\beta)$, we arrive at
\begin{align}
    \pi^{(q)}_{\beta}(z) = \frac{1}{Z_q(\beta)} \, \tpi_0(\vz)\exp_q \left\{\beta \cdot  \phi(\vz) \right\} \qquad
    Z_q(\beta) &:= \int  \tpi_0(\vz) \exp_q \left\{ \beta \cdot \phi(\vz)\right\} \, d\vz .\label{eq:lnq_fam_form2}
\end{align}

\subsection{Normalization in q-Exponential Families}\label{app:normalization}
The $q$-exponential family can also be written using the $q$-free energy $\psi_q(\theta)$ for normalization \cite{amari2011q, naudts2011generalised},
\begin{align}
    \pi_{\theta}^{(q)}(\z)  &= \pi_0(z) \, \exp_q \big\{ \theta \cdot \phi(\vz) - \psi_q(\theta ) \big\} \, . \label{eq:qexp_fam_qf} 
\end{align}
However, since $\exp_q \{ x + y \} = \exp_q \{ y \} \cdot \exp_q \{ \frac{x}{1 + (1-q) y}  \}$ (see \cite{suyari2020advantages} or App. \ref{app:q_sum_product} below) instead of $\exp \{ x + y \} = \exp \{ x \} \cdot \exp \{ y \} $ for the standard exponential, we can not easily move between these ways of writing the $q$-family \cite{matsuzoe2019normalization}.

Mirroring the derivations of \citet{naudts2011generalised} pg. 108, we can rewrite \eqref{eq:qexp_fam_qf} using the above identity for $\exp_q \{ x+y\}$, as
\begin{align}
\pi^{(q)}_{\theta}(\vz) &= \pi_0(\vz) \, \exp_q \{ \theta \cdot \phi(\vz) - \psi_q(\theta) \} \label{eq:normalization1} \\
&= \pi_0(\vz) \, \exp_q \{ - \psi_q(\theta) \} \exp_q \big\{ \frac{\theta \cdot \phi(\vz)}{1+(1-q)(-\psi_q(\theta))} \big \} \label{eq:normalization2}
\end{align}
Our goal is to express $\pi^{(q)}_{\theta}(\vz)$ using a normalization constant $Z^{(q)}_\beta$ instead of the $q$-free energy $\psi_q(\theta)$.  While the exponential family allows us to freely move between $\psi(\theta)$ and $\log Z_{\theta}$, we must adjust the natural parameters (from $\theta$ to $\beta$) in the $q$-exponential case.   Defining
\begin{align}
 \beta &= \frac{\theta}{1+(1-q)(-\psi_q(\theta))} \\
Z^{(q)}_\beta &= \frac{1}{\exp_q \{-\psi_q(\theta) \}} 
\end{align}
we can obtain a new parameterization of the $q$-exponential family, using parameters $\beta$ and multiplicative normalization constant $Z^{(q)}_\beta$,
\begin{align}
 \pi^{(q)}_{\beta}(\vz) &=  \frac{1}{Z^{(q)}_\beta} \pi_0(z) \, \exp_{q} \{ \beta \cdot \phi(\vz) \} \\
 &= \pi_0(z) \, \exp_q \big\{ \theta \cdot \phi(\vz) - \psi_q(\theta ) \big\} = \pi^{(q)}_{\theta}(\vz)  \, .
\end{align}
See \citet{matsuzoe2019normalization}, \citet{suyari2020advantages}, and \citet{naudts2011generalised} for more detailed discussion of normalization in deformed exponential families.

\section{Minimizing $\alpha$-divergences}\label{app:alpha_integration}
\citet{amari2007integration} shows that the $\alpha$ power mean $\pi^{(\alpha)}_{\beta}$ minimizes the expected divergence to a single distribution, for \textit{normalized} measures and $\alpha = 2q-1$.   We repeat similar derivations but for the case of unnormalized endpoints $\{\tpi_i\}$ and $\tilde{r}(\vz)$
\begin{align}
\tpi_{\alpha}(\vz) &= \argmin \limits_{\tilde{r}(\vz)} \sum \limits_{i=1}^{N} w_i \, D_{\alpha}[\tpi_i (\vz) : \tilde{r}(\vz) ] \label{eq:amari_vrep} \\
\text{where} \quad \tpi_{\alpha}(\vz) &= \big( \sum \limits_{i=1}^{N} w_i \, \tpi_i (\vz)^{\frac{1-\alpha}{2}}     \big)^{\frac{2}{1-\alpha}} \label{eq:alpha_mix_app}
\end{align}
\begin{proof}
\begin{align}
\frac{d}{d\tilde{r}} \sum \limits_{i=1}^{N} w_i \, D_{\alpha}[\tpi_i (\vz) : \tilde{r}(\vz) ] &= \frac{d}{d\tilde{r}} \frac{4}{1-\alpha^2} \sum \limits_{i=1}^{N} w_i \big( -\int \tpi_i(\vz)^{\frac{1-\alpha}{2}}\, \tilde{r}(\vz)^{\frac{1+\alpha}{2}} d\vz  +  \frac{1+\alpha}{2} \int \tilde{r}(\vz) d\vz \big) \\
0 &=\frac{4}{1-\alpha^2} \big( -{\frac{1+\alpha}{2}} \sum \limits_{i=1}^{N} w_i  \, \tpi_i(\vz)^{\frac{1-\alpha}{2}}\, \tilde{r}(\vz)^{\frac{1+\alpha}{2}-1} + \frac{1+\alpha}{2} \big) \\
- \frac{2}{1-\alpha} &= -\frac{2}{1-\alpha} \sum \limits_{i=1}^{N} w_i  \, \tpi_i(\vz)^{\frac{1-\alpha}{2}}  \tilde{r}(\vz)^{-\frac{1-\alpha}{2}} \\
 \tilde{r}(\vz)^{\frac{1-\alpha}{2}} &= \sum \limits_{i=1}^{N} w_i  \, \tpi_i(\vz)^{\frac{1-\alpha}{2}} \\
  \tilde{r}(\vz) &= \bigg(\sum \limits_{i=1}^{N} w_i  \, \tpi_i(\vz)^{\frac{1-\alpha}{2}} \bigg)^{\frac{2}{1-\alpha}}
\end{align}
\end{proof}
This result is similar to a general result about Bregman divergences in \citet{Banerjee2005} Prop. 1. although $D_{\alpha}$ is not a Bregman divergence over normalized distributions.

\subsection{Arithmetic Mean ($q=0$)}\label{app:mixture_path}
\newcommand{\pa}{\pi_0}
\newcommand{\pb}{\pi_1}
\newcommand{\opt}{r}
For normalized distributions, we note that the moment-averaging path from \citet{grosse2013annealing} is \textit{not} a special case of the $\alpha$-integration \cite{amari2007integration}.   While both minimize a convex combination of reverse \textsc{kl} divergences, \citet{grosse2013annealing} minimize within the constrained space of exponential families, 
 while \citet{amari2007integration} optimizes over \textit{all} normalized distributions.

More formally, consider minimizing the functional
\begin{align}
    J[\opt] &= (1 - \beta)\int \pa(z) \log \frac{\pa(z)}{\opt(z)} dz + \beta \int \pb(z) \log \frac{\pb(z)}{\opt(z)} dz \\
         &= \text{const} - \int \left[(1 - \beta) \pa(z) + \beta \pb(z) \right] \log \opt(z) dz \label{eq:functional}
\end{align}
We will show how \citet{grosse2013annealing} and \citet{amari2007integration} minimize \eqref{eq:functional}.

\paragraph{Solution within Exponential Family}
\citet{grosse2013annealing} constrains $\opt(z) = \frac{1}{Z(\theta)} h(z) \exp (\theta^T g(z))$ to be a (minimal) exponential family model and minimizes \eqref{eq:functional} w.r.t $\opt$'s  natural parameters $\theta$ (cf. \cite{grosse2013annealing} Appendix 2.2):
\begin{align}
    \theta^*_i &= \argmin_\theta J(\theta) \\
    &= \argmin_\theta \left(- \int \left[(1 - \beta) \pa(z) + \beta \pb(z) \right] \left[ \log h(z) + \theta^T g(z) - \log Z(\theta) \right] dz \right)\\
    &= \argmin_\theta \left(\log Z(\theta) - \int \left[(1 - \beta) \pa(z) + \beta \pb(z) \right] \theta^T g(z) dz  + \text{const}\right)
\end{align}
where the last line follows because $\pa(z)$ and $\pb(z)$ are assumed to be correctly normalized. Then to arrive at the moment averaging path, we compute the partials $\frac{\partial J(\theta)}{\partial \theta_i}$ and set to zero:
\begin{align}
    \frac{\partial J(\theta)}{\partial \theta_i} &= \E_{\opt}[g_i(z)] - (1 - \beta)\E_{\pa}[g_i(z)] - \beta \E_{\pb}[g_i(z)] = 0 \\
    \E_{\opt}[g_i(z)] &= (1 - \beta)\E_{\pa}[g_i(z)] - \beta \E_{\pb}[g_i(z)]
\end{align}
where we have used the exponential family identity $\frac{\partial \log Z(\theta)}{\partial \theta_i} = \E_{\opt_{\theta}}[g_i(z)]$ in the first line.

\paragraph{General Solution}
Instead of optimizing in the space of minimal exponential families, \citet{amari2007integration} instead adds a Lagrange multiplier to \eqref{eq:functional} and optimizes $\opt$ directly (cf. \cite{amari2007integration} eq. 5.1 - 5.12)
\begin{align}
    {\opt}^* &= \argmin_{\opt} J'[\opt] \\
    &= \argmin_{\opt} J[\opt] + \lambda \left(1 - \int \opt(z) dz\right) \label{eq:lagrange}
\end{align}
\cref{eq:lagrange} can be minimized using the Euler-Lagrange equations or using the identity
\begin{align}
    \frac{\delta f(x)}{\delta f(x')} = \delta(x - x')\label{eq:delta_eq}
\end{align}
from \cite{meng2004introduction}. We compute the functional derivative of $J'[\opt]$ using \eqref{eq:delta_eq} and solve for $r$:
\begin{align}
    \frac{\delta J'[\opt]}{\delta \opt(z)}=&- \int \big[(1 - \beta) \pa(z') + \beta \pb(z') \big] \frac{1}{\opt(z')} \frac{\delta \opt(z')}{\delta \opt(z)} dz' - \lambda \int \frac{\delta \opt(z')}{\delta \opt(z)} dz' \\
    =&- \int \big[(1 - \beta) \pa(z') + \beta \pb(z') \big] \frac{1}{\opt(z')} \delta(z - z') dz' - \lambda \int \delta(z - z') dz' \\
    =&- \big[(1 - \beta)\pa(z) + \beta \pb(z)\big] \frac{1}{\opt(z)} - \lambda = 0
\end{align}
Therefore
\begin{align}
    \opt(z) \propto \big[(1 - \beta)\pa(z) + \beta \pb(z)\big],
\end{align}
which corresponds to our $q$-path at $q=0$, or $\alpha = -1$ in \citet{amari2007integration}.  Thus, while both \citet{amari2007integration} and \citet{grosse2013annealing} start with the same objective, they arrive at different optimum because they optimize over different spaces.

\section{Sum and Product Identities for $q$-Exponentials}\label{app:q_sum_product}
In this section, we prove two lemmas which are useful for manipulation expressions involving $q$-exponentials, for example in moving between \cref{eq:normalization1} and \cref{eq:normalization2} in either direction.
\begin{lemma}
    Sum identity
    \begin{align}
        \exp_q\left(\sum_{n=1}^N x_n\right) = \prod_{n=1}^{N} \exp_q \left(\frac{x_n}{1 + (1 - q)\sum_{i=1}^{n-1}x_i} \right)\label{eq:q_exp_sum}
    \end{align}
    \label{lemma:q_exp_sum}
\end{lemma}
\begin{lemma}
    Product identity
    \begin{align}
        \prod_{n=1}^N \exp_q(x_n) = \exp_q\left(\sum_{n=1}^{N}x_n \cdot \prod_{i=1}^{n-1} \left(1 + (1 - q)x_i\right)\right)\label{eq:q_exp_prod}
    \end{align}
    \label{lemma:q_exp_prod}
\end{lemma}

\subsection{Proof of \cref{lemma:q_exp_sum}}
\begin{proof}
    We prove by induction. The base case ($N=1$) is satisfied using the convention $\sum_{i=a}^bx_i = 0$ if $b < a$ so that the denominator on the \textsc{rhs} of \cref{eq:q_exp_sum} is $1$. Assuming \cref{eq:q_exp_sum} holds for $N$,
    \begin{align}
        \exp_q\left(\sum_{n=1}^{N+1} x_n\right) &= \left[ 1 + (1-q) \sum_{n=1}^{N+1} x_n \right]_{+}^{1/(1-q)} \\
                                                &= \left[ 1 + (1-q) \left(\sum_{n=1}^{N} x_n\right) + (1-q)x_{N+1} \right]_{+}^{1/(1-q)} \\
                                                &= \left[\left( 1 + (1-q) \sum_{n=1}^{N} x_n \right) \left(1 + (1-q)\frac{x_{N+1}}{1 + (1-q) \sum_{n=1}^{N} x_n}\right) \right]_{+}^{1/(1-q)} \\
                                                &= \exp_q\left(\sum_{n=1}^N x_n\right) \exp_q \left(\frac{x_{N+1}}{1 + (1-q) \sum_{n=1}^{N} x_n} \right)\\
                                                &= \prod_{n=1}^{N+1} \exp_q \left(\frac{x_n}{1 + (1-q)\sum_{i=1}^{n-1}x_i} \right) \text{(using the inductive hypothesis)}
    \end{align}
\end{proof}
\subsection{Proof of \cref{lemma:q_exp_prod}}
\begin{proof}
    We prove by induction. The base case ($N=1$) is satisfied using the convention $\prod_{i=a}^bx_i = 1$ if $b < a$. Assuming \cref{eq:q_exp_prod} holds for $N$, we will show the $N+1$ case. To simplify notation we define $y_N:=\sum_{n=1}^{N}x_n \cdot \prod_{i=1}^{n-1} \left(1 + = (1 - q)x_i\right)$. Then, 
\begin{align}
    \prod_{n=1}^{N+1} \exp_q(x_n) &= \exp_q(x_{1})\left(\prod_{n=2}^{N+1}\exp_q(x_n)\right)\\
    &= \exp_q(x_{0})\left(\prod_{n=1}^{N}\exp_q(x_n)\right) & \hspace*{-.5cm} \text{(reindex $n \to n - 1)$} \nonumber \\
    &=\exp_q(x_{0})\exp_q(y_N) & \hspace*{-.5cm} \text{(inductive hypothesis)} \nonumber \\
    &= \bigg[\left(1 + (1-q) \cdot x_{0}\right)\left(1 + (1-q) \cdot y_N\right) \bigg]_{+}^{1/(1-q)}\\
    &= \bigg[1 + (1-q) \cdot x_{0} + \big(1 + (1-q) \cdot x_{0} \big)(1-q) \cdot y_N \bigg]_{+}^{1/(1-q)}\\
    &= \bigg[1 + (1-q) \bigg(x_{0} + \big(1 + (1-q) \cdot x_{0} \big)y_N\bigg) \bigg]_{+}^{1/(1-q)}\\
    &= \exp_q \left(x_{0} + \big(1 + (1-q) \cdot x_{0} \big)y_N\right)
\end{align}
Next we use the definition of $y_N$ and rearrange
\begin{align}
    &= \exp_q \left(x_{0} + \big(1 + (1-q) \cdot x_{0} \big)\left(x_1 + x_2(1 + (1-q) \cdot x_1) + ... + x_N \cdot \prod_{i=1}^{N-1}(1 + (1-q) \cdot x_i)\right)\right) \nonumber\\
    &= \exp_q\left(\sum_{n=0}^{N}x_n \cdot \prod_{i=1}^{n-1} \left(1 + (1-q) x_i\right)\right).
\end{align}
Then reindexing $n \to n + 1$ establishes
\begin{align}
    \prod_{n=1}^{N+1} \exp_q(x_n) = \exp_q\left(\sum_{n=1}^{N+1}x_n \cdot \prod_{i=1}^{n-1} \left(1 + (1-q)x_i\right)\right).
\end{align}
\end{proof}

\newcommand{\qqq}{{q}}
\newcommand{\qqexp}{{1-q}}
\newcommand{\qqexpconj}{{q}}
\newcommand{\qqinv}{{{\frac{1}{1-q}}}}
\newcommand{\bmu}{{\mathbf{\mu}}}
\newcommand{\bSig}{{\mathbf{\Sigma}}}

\section{Annealing between Student-$t$ Distributions}\label{app:additional} 


\subsection{Student-$t$ Distributions and $q$-Exponential Family}
The Student-$t$ distribution appears in hypothesis testing with finite samples, under the assumption that the sample mean follows a Gaussian distribution.   In particular, the degrees of freedom parameter $\nu = n-1$ can be shown to correspond to an order of the $\qqq$-exponential family with $\nu = (3-\qqq) / (\qqq-1)$ (in 1-d), so that the choice of $\qqq$ is linked to the amount of data observed.

We can first write the multivariate Student-$t$ density, specified by a mean vector $\mu$, covariance $\bSig$, and degrees of freedom parameter $\nu$, in $d$ dimensions, as
\begin{align}
    t_{\nu}(\vx | \bmu, \bSig) = \frac{1}{Z(\nu, \bSig)} \big[ 1 + \frac{1}{\nu} (x - \bmu)^T \bSig^{-1} (x-\bmu) \big]^{-\big(\frac{\nu + d }{2}\big)} \label{eq:student}
\end{align}
where $Z(\nu, \bSig) = \Gamma(\frac{\nu+d}{2})/\Gamma(\frac{\nu}{2}) \cdot |\bSig|^{-1/2} \nu^{-\frac{d}{2}} \pi^{-\frac{d}{2}}$.  Note that  $\nu > 0$, so that we only have positive values raised to the $-(\nu+d)/2$ power, and the density is defined on the real line.   

The power function in \eqref{eq:student} is already reminiscent of the $\qqq$-exponential, while we have first and second moment sufficient statistics as in the Gaussian case.  We can solve for the exponent, or order parameter $q$, that corresponds to $-(\nu+d)/2$ using $-\big(\frac{\nu + d }{2}\big) = \frac{1}{1-\qqq}$.  This results in the relations
\begin{align}
 \nu = \frac{d - d \qqq +2}{\qqq - 1} \qquad \text{or} \qquad \qqq = \frac{\nu+d+2}{\nu+d} 
\end{align}
We can also rewrite the $\nu^{-1} \, (x - \bmu)^T \bSig^{-1} (x-\bmu) $ using natural parameters corresponding to $\{x, x^2\}$ sufficient statistics as in the Gaussian case (see, e.g. \citet{matsuzoe2015deformed} Example 4).

Note that the Student-$t$ distribution has heavier tails than a standard Gaussian, and reduces to a multivariate Gaussian as $\qqq \rightarrow 1$ and $\exp_{\qqq}(u) \rightarrow \exp(u)$.  This corresponds to observing $n\rightarrow \infty$ samples, so that the sample mean and variance approach the ground truth \cite{murphy2007conjugate}. 

\begin{figure}[t]
    \centering
    \subfigure[$q=0$] {\includegraphics[trim={0 1.2cm 0 1cm},clip,width=0.19\textwidth]{sections/figs/gif/ridge_01.pdf}}
    \subfigure[$q=0.5$]{\includegraphics[trim={0 1.2cm 0 1cm},clip,width=0.19\textwidth]{sections/figs/gif/ridge_09.pdf}}
    \subfigure[$q=0.9$]{\includegraphics[trim={0 1.2cm 0 1cm},clip,width=0.2\textwidth]{sections/figs/gif/ridge_18.pdf}}
    \subfigure[$q=1$]{\includegraphics[trim={0 1.2cm 0 1cm},clip,width=0.19\textwidth]{sections/figs/gif/ridge_20.pdf}}
    \subfigure[$q=2$]{\includegraphics[trim={0 .1cm 0.25cm 0 },clip,width=0.19\textwidth]{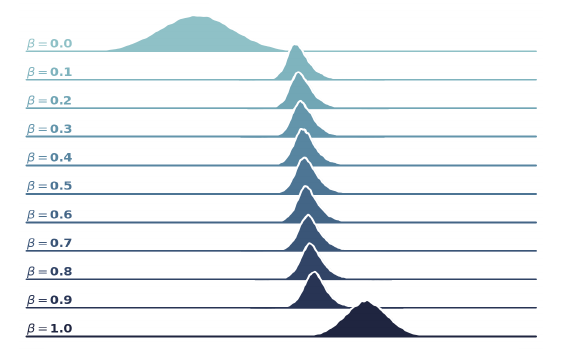}}
    \caption{Intermediate densities between $\mathcal{N}(-4, 3)$ and $\mathcal{N}(4,1)$ for various $q$-paths and 10 equally spaced $\beta$. The path approaches a mixture of Gaussians with  weight $\beta$ at $q=0$.  For the geometric mixture ($q=1$), intermediate $\pi_{\beta}$ stay within the exponential family since both $\pi_0$, $\pi_T$ are Gaussian.}
    \label{fig:alpha_path22} 
    \label{fig:gaussian_path} 
    \vspace*{-.15cm}
\end{figure}%
\begin{figure}[t]
    \centering
    \includegraphics[trim={0 0 0 0 },clip,width=0.99\textwidth]{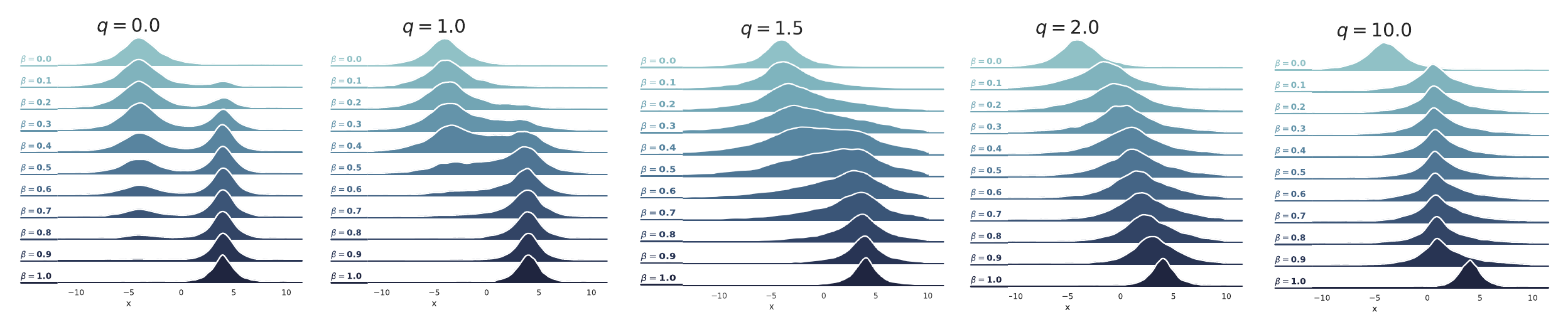}
    \caption{Intermediate densities between Student-$t$ distributions, $t_{\nu = 1}(-4, 3)$ and $t_{\nu = 1}(4,1)$ for various $q$-paths and 10 equally spaced $\beta$, 
    Note that $\nu=1$ corresponds to $q=2$, so that the $q=2$ path stays within the $q$-exponential family.\\[3ex]
    We provide code to reproduce experiments at \url{https://github.com/vmasrani/q\_paths}.
    }
    \label{fig:alpha_path} 
    \label{fig:student_path} 
    \vspace*{-.15cm}
\end{figure}

\subsection{Annealing between 1-d Student-$t$ Distributions}
Since the Student-$t$ family generalizes the Gaussian distribution to $q \neq 1$, we can run a similar experiment annealing between two Student-$t$ distributions.   We set $q=2$, which corresponds to $\nu = 1$ with $\nu = (3-\qqq) / (\qqq-1)$, and use the same mean and variance as the Gaussian example in Fig. \ref{fig:alpha_path22}, with $\pi_0(z) = t_{\nu=1}( -4, 3)$ and $\pi_1(z) = t_{\nu=1}( 4, 1)$.   

We visualize the results in Fig. \ref{fig:student_path}.   For this special case of both endpoint distributions within a parametric family, we can ensure that the $q=2$ path stays within the $q$-exponential family of Student-$t$ distributions.   We make a similar observation for the Gaussian case and $q=1$ in Fig. \ref{fig:gaussian_path}.   Comparing the $q=0.5$ and $q=0.9$ Gaussian path with the $q=1.0$ and $q=1.5$ path, we observe that mixing behavior appears to depend on the relation between the $q$-path parameter and the order of the $q$-exponential family of the endpoints.   

As $q \rightarrow \infty$, the power mean \eqref{eq:abstract_mean} approaches the $\min$ operation as $1-q \rightarrow -\infty$.    In the Gaussian case, we see that, even at $q=2$, intermediate densities for all $\beta$ appear to concentrate in regions of low density under both $\pi_0$ and $\pi_T$.   However, for the heavier-tailed Student-$t$ distributions, we must raise the $q$-path parameter significantly to observe similar behavior.

\newcommand{\base}{g}
\subsection{Endpoints within a Parametric Family}
If the two endpoints $\tpi_0, \tpi_T$ are within a $q$-exponential family, we can show that each intermediate distribution along the $q$-path of the same order is also within this $q$-family.  However, we cannot make such statements for general endpoint distributions or members of  different $q$-exponential families.

\paragraph{Exponential Family Case}
We assume potentially vector valued parameters $\theta = \{ \theta\}_{i=1}^N$ with multiple sufficient statistics $\phi(\vz) = \{ \phi_i(\vz) \}_{i=1}^N$, with $\theta \cdot \phi(\vz) = \sum_{i=1}^N \theta_i \phi_i(\vz)$.
For a common base measure $\base(\vz)$, let $\tpi_0(\vz) = \base(\vz) \, \exp\{ \theta_0 \cdot \phi(\vz) \}$ and $\tpi_1(\vz) = \base(\vz) \, \exp \{ \theta_1 \cdot \phi(\vz) \}$.   Taking the geometric mixture,
\begin{align}
    \tpi_\beta(\vz) &= \exp \big\{ (1-\beta) \, \log \tpi_0(\vz) + \beta \, \log \tpi_T(\vz) \big\} \\
    &= \exp \big \{ \log \base(\vz) + (1-\beta) \, \theta_0 \cdot \phi(\vz) + \beta  \, \theta_1 \phi(\vz)  \big \} \\
    &= \base(\vz) \exp  \big \{ \big( (1-\beta) \, \theta_0 + \beta  \, \theta_1 \big) \cdot \phi(\vz)  \big \}
\end{align}
which, after normalization, will be a member of the exponential family with natural parameter $(1-\beta) \, \theta_0 + \beta  \, \theta_1$.

\paragraph{$q$-Exponential Family Case} For a common base measure $\base(\vz)$, let $\tpi_0(\vz) = \base(\vz) \, \exp_q \{ \theta_0 \cdot \phi(\vz) \}$ and $\tpi_1(\vz) = \base(\vz) \, \exp_q \{ \theta_1 \cdot \phi(\vz) \}$.   The $q$-path intermediate density becomes
\begin{align}
\tilde{\pi}^{(q)}_\beta(\vz) &= \big[ (1-\beta) \, \tpi_0(\vz)^{1-q} + \beta \, \tpi_T(\vz)^{1-q} \big]^{\frac{1}{1-q}} \\
&= \big[ (1-\beta) \, \base(\vz)^{1-q} \, \exp_q \{ \theta_0 \cdot \phi(\vz) \}^{1-q} + \beta \, \base(\vz)^{1-q} \,\exp_q \{ \theta_1 \cdot \phi(\vz) \} ^{1-q} \big]^{\frac{1}{1-q}} \\
&=  \bigg[ \base(\vz)^{1-q} \bigg ( (1-\beta) \,  \, [1 + (1-q)( \theta_0 \cdot \phi(\vz))]^{\frac{1}{1-q}1-q} + \beta  \, [1 + (1-q)( \theta_1 \cdot \phi(\vz))]^{\frac{1}{1-q} 1-q} \bigg) \bigg]^{\frac{1}{1-q}} \nonumber \\
&= \base(\vz) \bigg[ 1 + (1-q) \bigg( \big((1-\beta) \, \theta_0 + \beta  \, \theta_1 \big)  \cdot \phi(\vz) \bigg) \bigg]^{\frac{1}{1-q}} \\
&= \base(\vz) \exp_q \big\{  \big((1-\beta) \, \theta_0 + \beta  \, \theta_1 \big) \cdot \phi(\vz) \big\} 
\end{align}
which has the form of an unnormalized $q$-exponential family density with parameter $(1-\beta) \, \theta_0 + \beta  \, \theta_1$.

\end{document}